\documentclass[pmlr,twocolumn, wcp]{jmlr} 
\usepackage[utf8]{inputenc} 
\usepackage[T1]{fontenc}    
\usepackage{hyperref}       
\usepackage{url}            
\usepackage{float}
\usepackage{booktabs}       
\usepackage{array}
\usepackage{amsfonts}       
\usepackage{nicefrac}       
\usepackage{microtype}      
\usepackage{graphicx}
\usepackage{amsmath}
\usepackage{appendix}
\jmlrworkshop{Machine Learning for Health (ML4H) 2020} 

\usepackage{color}
\definecolor{orange}{rgb}{1,0.5,0}

\title{sEMG Gesture Recognition with a Simple Model of Attention}
\author{%
  \Name{David Josephs} \Email{josephsd@smu.edu}\\
    \Name{Carson Drake} \Email{drakec@smu.edu}\\
    \Name{Andy Heroy} \Email{aheroy@smu.edu}\\
    \Name{John Santerre} \Email{jsanterre@smu.edu}\\
    \addr{Southern Methodist University, Dallas, TX 75205}
}

\begin{document}
\maketitle
\begin{abstract}

 Myoelectric control is one of the leading areas of research in the field of robotic prosthetics. We present our research in surface electromyography (sEMG) signal classification, where our simple and novel attention-based approach now leads the industry, universally beating more complex, state-of-the-art models. Our novel attention-based model achieves benchmark leading results on multiple industry-standard datasets including 53 finger, wrist, and grasping motions, improving over both sophisticated signal processing and CNN-based approaches. Our strong results with a straightforward model also indicate that sEMG represents a promising avenue for future machine learning research, with applications not only in prosthetics, but also in other important areas, such as diagnosis and prognostication of neurodegenerative diseases, computationally mediated surgeries, and advanced robotic control. We reinforce this suggestion with extensive ablative studies, demonstrating that a neural network can easily extract higher order spatiotemporal features from noisy sEMG data collected by affordable, consumer-grade sensors. 

\end{abstract}
\section{Introduction}

    Electrophysiological studies of the nervous system are a core area of research in clinical neurophysiology, where scientists attempt to link electrical signals from the body to their physical manifestations. These studies include measuring brain waves (electroencephalography), comparison of sensory stimuli to electrical signals in the central nervous system (evoked potential), and the measure of electrical signals in skeletal muscles (electromyography), which is the focus of this paper.

When the nervous system uses signals to communicate with a skeletal muscle, the myocytes (muscle cells) contract, causing physical motion and electrical excitation. By measuring these changes in electric potential in myocytes, we correlate signals to intended action. Such an approach allows for opportunities ranging from quantifying physical veracity to diagnosing neurodegenerative diseases. An example of the latter can be found in \citet{graves}, where electromyographic (EMG) signals were used to classify Multiple Sclerosis patients from healthy control subjects with 82\% accuracy.

In this paper, we choose to focus on surface electromyography (sEMG) classification with low-frequency data collected from cheap sensors, for use in robotic prosthetics. Amputations typically are more present in impoverished communities \citep{arya2018race, mackenzie2007health, mcginigle2014living}; therefore, we believe lowering the cost of access to high-grade prosthetics is of utmost importance. 

Current state-of-the-art myoelectric prosthetic limbs are capable of detecting and performing between 7 and 18 gestures \citep{myohandpro}. While there is great utility in such analysis, there remains considerable room for improvement. Our focus is on classifying signals used to dictate the motion of the hand. 



Our key contribution utilizing an attention mechanism to transform the data for consumption by a relatively shallow neural network.  Our work indicates that the key barrier holding back sEMG analysis is not limitations of interaction between the channels of signal (which might require a more complicated network structure) but rather the process by which we work with the time aspect of the classification problem.  Additionally,  we extend this simple model to show that attention, coupled with a simple neural network, can produce research-leading results against standard benchmarks, both NinaPro DB4 and DB5.  To our knowledge, models based purely on attention have never been shown to be effective tools for sEMG analysis, and our work represents an extension of \citet{aaai} for the discrete classification case. 

Specifically, we propose a  feed-forward attention-based architecture for sEMG recognition and myoelectric control. Our simple attention mechanism, in conjunction with feed-forward networks for time series classification, represents our main contribution. Using primarily the NinaPro DB5, as well as DB4, we demonstrate that a lightweight application of attention to sEMG data allows for benchmark leading results. 

%

\section{Background}
\subsection{Medical Background}

Electromyography is the measure of electrical activity generated from skeletal muscles.  We are particularly interested in brain mediated activity, where an electrical potential can be measured along the pathway of a signal originating in the central nervous system and transmitted to the muscular system.  The motor control process can be modeled as an axon, which delivers a signal to myocytes (muscle cells). The electrical signal causes the internal fibers of the muscle to contract to generate movement. The signals generated by these contractions, EMG signals, are measured by quantifying the difference in electrical potential between the inner and outer membranes of a myocyte. 
This is measured either through internal electrodes or electrodes on the surface of the skin (sEMG). We are particularly interested in sEMG, as although it is not as powerful as EMG, it is more practical and scalable (ready to be mass distributed) and less disruptive. These signal measurements are aggregates of several individual motor units. Though the sEMG signal has increased noise compared to intramuscular EMG, it presents an unintrusive and easily configurable means of measuring intent \citep{hussain2020intent}.  This is vital, as it allows an amputee whose limb does not biologically exist to make use of these signals still.

\subsection{Deep Learning for sEMG Classification} 

Much of the previous successful work in sEMG classification utilizes convolutional and recurrent neural networks to learn time, frequency, and time-frequency domain features. \citet{shen19} utilize all three of these domain features in conjunction with CNN's to classify the full NinaPro DB5 data set with 75\% accuracy. \citet{allard} builds off the work of \citet{atz_cnn}, using time and time-frequency domain transformations on raw sEMG signal and convolutional networks to obtain a balanced accuracy of 66.3\% over the wrist gesture subset (17 gestures + rest) of NinaPro DB5. \citet{mview} use multi-view CNNs on a set of classical time and frequency domain features and IMU data to score 91\% on the wrist and functional gestures of NinaPro DB5 . \citet{hu2018novel} looked to build a more robust representation of the time features of this data set using several CNN-RNN combinations, combined using attention mechanisms. 
\citet{rahimian2020xceptiontime} takes a novel approach to solve the time domain information problem, using a new separable depth wise convolutional architecture.
Parallel to this, \citet{mview, ketyko2019domain}, and \citet{Du_Sensors_2017} utilize domain adaptation techniques to help sEMG classifiers generalize not only across time, but across people, while \citet{allard, zhai2017self}, and \citet{zia2018multiday} also examine day-to-day calibration techniques, for practical use in amputees. 

One of the large problems in this space is determining how to represent the data in a manner in which neural networks can learn both spatial and temporal features. The literature indicates that there is a trade off between the complexity of the network and the complexity of preprocessing and feature extraction mechanisms \citep{hu2018novel, rahimian2020xceptiontime}.  We aim to show that with a simple architecture, minimal preprocessing, and no feature extraction, we can accurately and efficiently capture the relationship between sEMG and motion.

Accordingly, we explore adapting the language modeling notion of attention for sEMG analysis. We show that a feed-forward simplification of attention is capable of learning a time-domain representation of multichannel sEMG data. This approach manages to avoid the training and gradient issues of RNN's and the difficulty of data representation met by CNNs, performing well on near raw sEMG signals. The simplified attention mechanism discussed in \citet{raffel2015feedforward} is expanded upon for use with purely feed-forward networks. \citet{raffel2015feedforward} propose two approximations of a simple attention function, a simple sum across time as well as: 
\begin{equation}
\label{rafeq}
\alpha_t = \mathrm{softmax}\left(\mathrm{tanh}\left(W_{hc}h_t + b_t \right)\right).\end{equation}

Where $\alpha_t$ is the attention score at timestep $t$, $h_t$ is the output of the previous layer of the network, and $W_{hc}$ and $b_t$ are learned weights and biases. Using toy problems, \citet{raffel2015feedforward} demonstrate that feed-forward networks can use this mechanism to model sequences in which order is not important.

\section{Modeling}
\label{mod}

Our attention-based model architecture, shown in \autoref{fig:my_label}, consists of thre parts: the expansion and attention mechanism, and a classifier network. The data is fed into an expansion layer, which learns to expand the input data from a window of $C$ channels and $T$ timesteps to a matrix of $T$ timesteps and 128 channels. As seen in \autoref{tab:bottom}, the nature of this layer, fully connected or convolutional, is not important. This matrix is then processed by the feed-forward attention mechanism seen in \autoref{fig:my_label}, resulting in a feature vector of 128 values.  This feature vector is then fed into the classifier network consisting of two 500 node fully connected layers and one 2000 node fully connected layer as seen in \autoref{fig:my_label}. All parts of the classifier are separated by dropout connections.

Initially, we followed a CNN based approach in conjunction with attention. This had little effect on accuracy relative to a simple fully connected layer with attention. We include the CNN specifications here for clarity to further reinforce the value of the attention layer alone.  We used a single, 1-dimensional convolutional layer with a kernel size of 3 and same-padding at the bottom of the network in conjunction with layer normalization between each layer.
\begin{figure*}[t]
    \centering
    \includegraphics[width=0.95\textwidth]{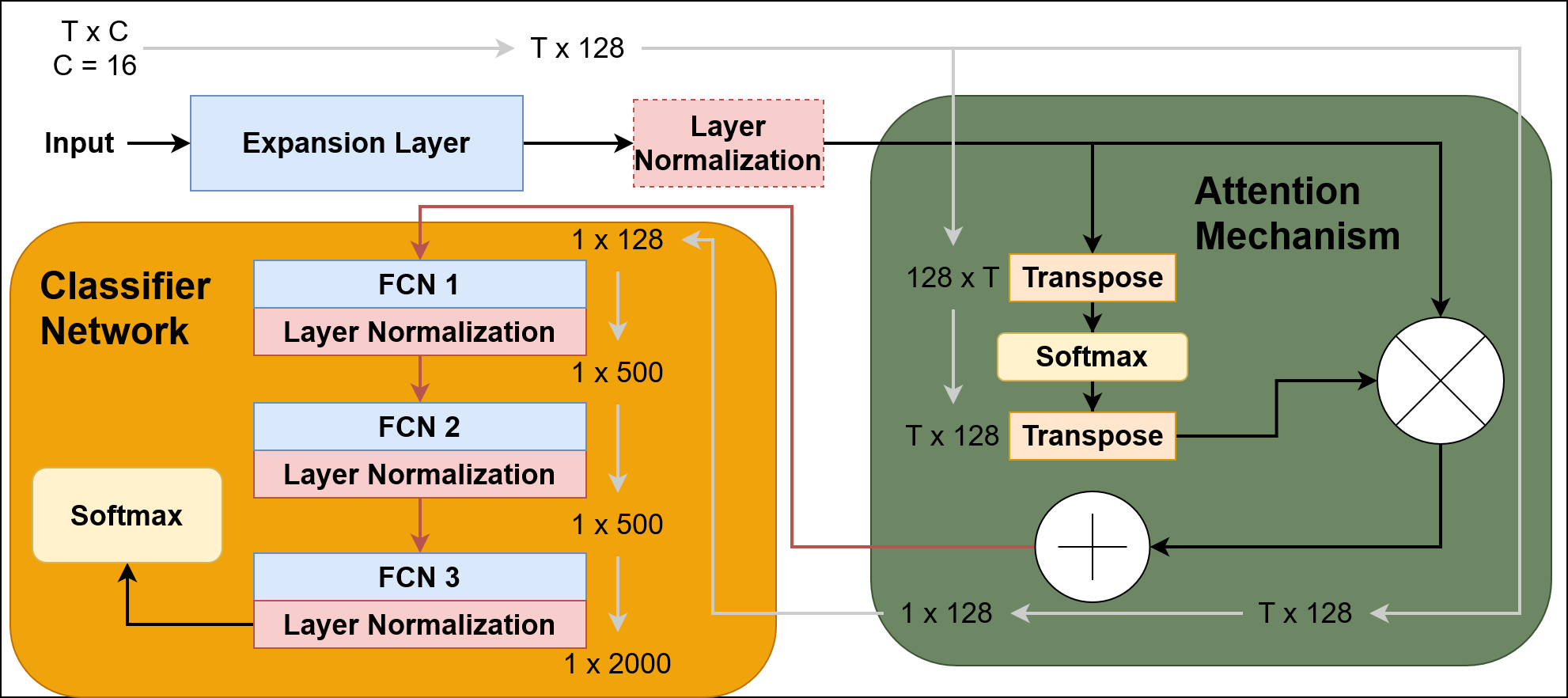}
    \caption{A diagram of our simple attention-based model. Note that red connections indicate dropout, and layer normalization is indicated as it contributed to our best results. The dimensions of a single window of input before and after each layer are shown, connected by gray arrows. FCN stands for ``fully connected network''.}
    \label{fig:my_label}
\end{figure*}


\subsection{Attention Mechanism}
The attention mechanism works as follows: First time series data processed by the previous layer, $h \,\epsilon\,\mathbb{R}^{T \times C}$,  where $T$ represents timesteps and $C$ represents features or channels, is inputted. $h$ is transposed into an $C \times T$ matrix. In $h^T$ an observation at a single row represents all timesteps of that feature or channel. For example, row $h^T_i$ represents the time series produced by the channel $C = i$. This matrix is then fed in row by row to a standard, fully connected layer using the softmax activation function, resulting in a $ C \times T$  matrix, $\alpha$.  This can be expressed as:

\begin{equation}
    \alpha = \mathrm{softmax}(W_{ht}h^T + b_{t}).
\end{equation}

Where $\alpha\,\epsilon\,\mathbb{R}^{C \times T}$ is an attention matrix and $W_{ht}\, \epsilon \,\mathbb{R}^{C \times T}, \, b\,\epsilon\,\mathbb{R}$ are learned weights and biases across time. For clarity, the entry $\alpha_{ij}$ is the attention score for timestep $j$ at channel $i$. Thus, $\alpha^T$ represents a matrix containing a learned temporal mask for each channel. By taking the Hadamard product of $\alpha^T$ and $h$, and summing across time, we calculate a vector $c \,\epsilon\,\mathbb{R}^{1\times C}$,  analogous to a context vector.
\begin{align}
g =  \alpha^{T} \circ h, \, g \, \epsilon \, \mathbb{R}^{T \times C} \\
c = \sum_{t=1}^T g, \, c \,\epsilon\,\mathbb{R}^{1 \times C}
\end{align} 

This naive attention mechanism proves surprisingly capable of learning expressive representations of the multichannel signal and can serve as a strong baseline model for future sEMG research. 



\subsection{Training and Implementation Details}
Each model was trained for 55 epochs with a batch size of 128, and a learning rate annealed from $1*10^{-3}$ to $1*10^{-5}$. We chose 55 epochs because after 45-55 epochs, the model did not improve any further.

Rather than using the Rectified Linear Unit (ReLU) activation function, all layers use the ``Mish'' activation function, proposed by \citet{mish}, with the idea of reaching a good optima in fewer iterations. Instead of optimizing the network with Adam or SGD, the model was optimized using the Ranger optimizer. The Ranger optimizer consists of two components: Rectified Adam (RAdam) and Lookahead. The RAdam algorithm represents an improvement over the Adam optimizer in that it does not require a ``warm up period''. This allows the model to be trained to an optima much faster \citep{radam}. The Lookahead algorithm works in conjunction with a primary optimizer. The primary optimizer calculates weights as it normally does, and then the Lookahead optimizer explores the loss landscape near the calculated weights. This allows for even faster convergence to an optima \citep{ranger2}. The combination of Lookahead and RAdam is the Ranger optimizer used in this paper \citep{ranger1}.  The learning rate follows a delayed cosine annealing schedule. For the first 5 epochs, the model trains at a high learning rate, and is subsequently annealed over the course of 50 epochs to a low learning rate. This training process takes between 3 and 4 hours on a single NVIDIA Tesla V100, provided by the Southern Methodist University Center for Research Computing. For regularization, dropout with a rate of 0.36 is applied to all layers in the classification network.

Large class imbalance presents a significant issue for laboratory-collected sEMG data. Most of the time, the hand is at rest, as each repetition of each gesture is followed by 3 seconds of rest. This is analogous to the real world, as we do not constantly move our hands; however, this complicates the classification task. In NinaPro DB5, there is over 30 times as much rest as any of the other 52 gestures. Several measures were taken to combat this. First, instead of training with standard cross-entropy, all models were trained using the focal loss function from computer vision, which weights easy to classify examples (rest) less than difficult examples when calculating the gradients. Secondarily, a novel data augmentation scheme was implemented, adding noise with a spectrum of Signal-To-Noise Ratios. For each window passed in, a random SNR between 1 and 30 is selected, and SNR-preserving noise is calculated and added to the window, see Appendix \ref{snr} for details. Due to the strong imbalance in the data, rest is not augmented. This is analogous to oversampling the minority class with synthetic data \citep{johnson2019survey}. Between 30 and 40 epochs, the model will have seen about the same number of unique samples per non-rest class as resting class samples. We propose using the Matthews Correlation Coefficient and Class-balanced accuracy as more appropriate evaluation metrics for future research.

\section{Data}
The MYO arm band \citep{myo} is a low-cost, \$120, low-frequency sEMG device. While other research grade devices achieve higher sampling frequencies, they are also very dependent on placement configuration and operator calibration. A critical requirement for a consumer grade sensor setup is ease of configuration, this is one advantage of consumer grade hardware, because exact positioning of the electrodes cannot be assured, the algorithmic approach taken must be more robust.  Thus choosing lower grade hardware puts greater emphasis on the necessity of algorithmic analysis and prepossessing data which is easily scale-able rather then expensive hardware and technical experience. 

The NinaPro \citep{atzori2015ninapro} project represents the largest data collection effort in the sEMG space, consisting of 10 large databases collected from both amputees and intact subjects using various sensors. This paper's main benchmark is the NinaPro DB5, or the ``Double MYO'' dataset.  The NinaPro DB5 uses two MYO armbands, offset in both position and angle, collecting 16 channels of sEMG data, as well as triaxial accelerometer (Inertial Measurement Unit or IMU) data. Information on the test subjects such as arm circumference, gender, and age can be found in \citet{nina5}. Additionally, some analysis was done with the NinaPro DB4 dataset, which utilized the same experimental design but collected using 12 sensors \citep{nina5}.

For both DB5 and DB4, there are 53 unique gestures measured, collected over 10 subjects. The subjects perform six repetitions of each gesture, buffering each gesture repetition with three seconds of rest. The databases contain 12 fine finger gestures, 17 fine wrist gestures, 23 functional (grasping) gestures, and rest. This data looks like a 16 channel time series with approximately 600000 rows per subject. Recording 6 repetitions per gesture, each subject performed a total of 318 movements. In DB5, the signal is collected at 200 Hz, the sampling frequency of the MYO armband.  DB4 collects 12 channels of sEMG at 2 kHz using Cometa differential electrodes on the same experimental design, providing a good high-frequency comparison to DB5. 

\begin{figure}
    \centering
    \def\svgwidth{0.8\columnwidth}
    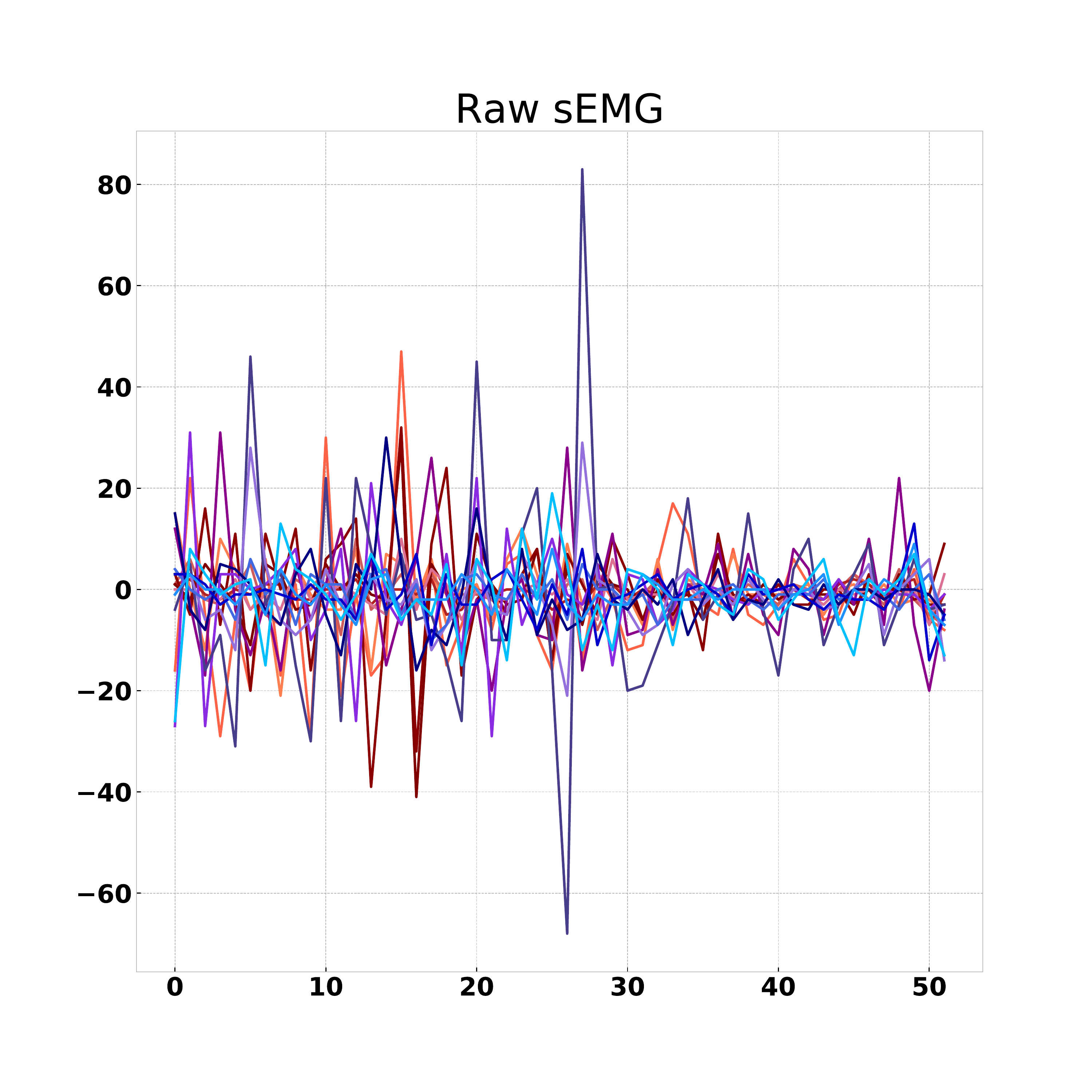
    \caption{Unprocessed sEMG data.}
\label{fig:preproc}
\end{figure}
\begin{figure}
\begin{center}
    
\def\svgwidth{0.8\columnwidth}
    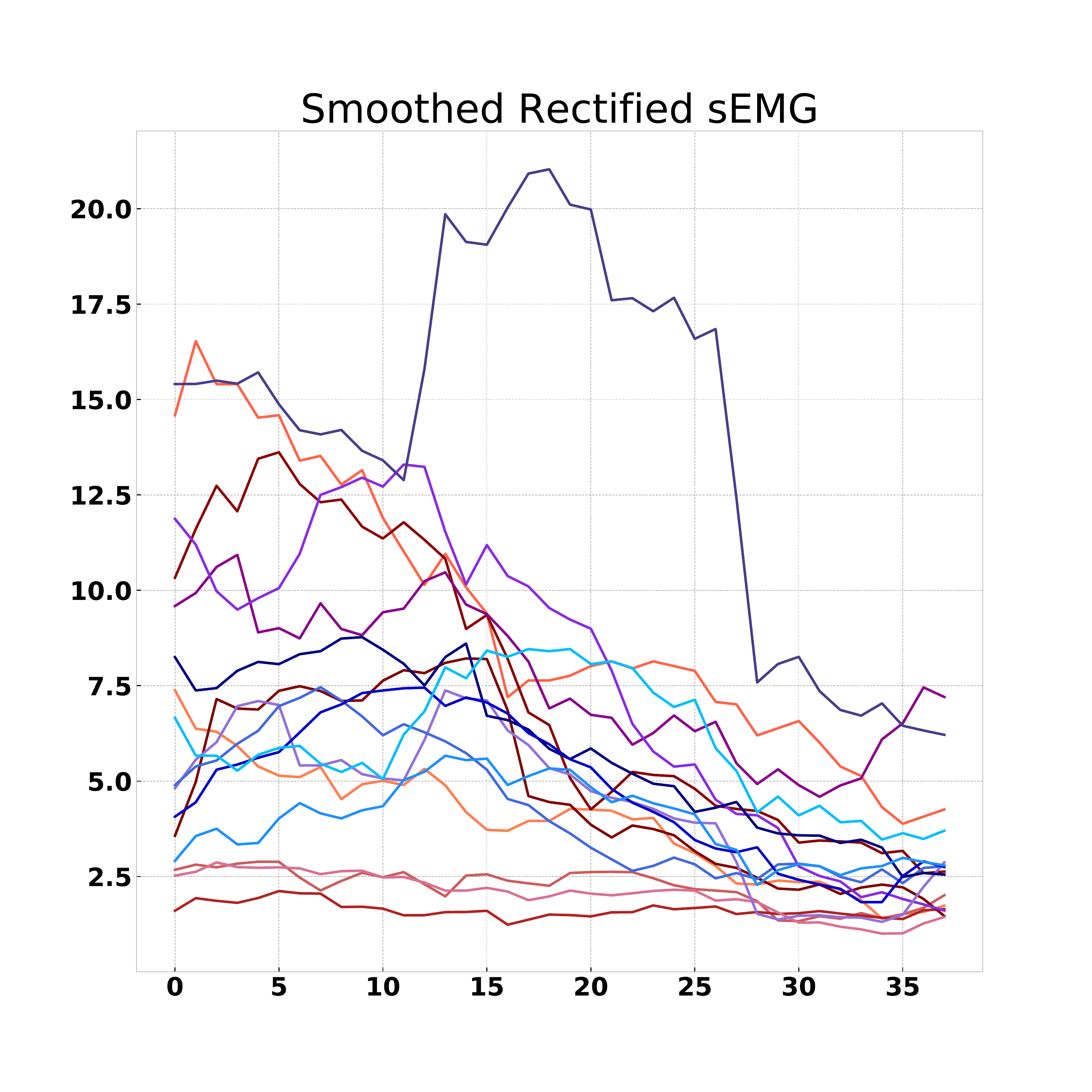
\end{center}
\caption{Processed sEMG data.}
\label{fig:proc}
\end{figure}

In \autoref{fig:preproc}, we see one window of raw sEMG data. The MYO sensor has already removed any powerline interference with a notch filter. Next, we see the data after rectification (absolute value), filtering, and smoothing in \autoref{fig:proc}. Next,
The data was processed as follows. First, following the precedent set by \citet{allard}, the raw data stream is broken into 260-millisecond windows, with 235 milliseconds of overlap. This window size was chosen in part due to the work of \citet{250ms} and \citet{300ms}, who determined that the absolute largest window size for use in a prosthetic limb lies somewhere between 250 and 300 milliseconds. Windows which contain multiple gestures (generally rest and another gesture) are labeled using the first gesture, meaning a window which is mostly rest at the beginning, and then a grabbing gesture will be classified as rest, and a window which is mostly a grabbing gesture followed by a rest is classified as a grabbing gesture. Windows containing multiple repetitions were discarded. One window of raw sEMG data can be seen in \autoref{fig:preproc}, the data is rectified, which exposes more information on the firing rate of the motor units \citep{rectif}, and artifacts are removed with a 20 Hz high-pass Butterworth filter. The order of these two steps is important, as the rectification alters the signal's power spectrum, and thus alters the results of the filter. Finally, we smooth the data with a moving average filter. This brings $T$, the number of timesteps in a window, down from $52$ to $38$ and $520$ to $381$ in DB5 and DB4 respectively. This can be seen in \autoref{fig:proc} The third repetition was used to tune parameters and the fifth as an unseen holdout set for all trials. The fifth repetition is commonly used in testing \citep{atz_cnn, mview, nina5}.

\section{Results}
\begin{figure}[h!]
    \centering
    \includegraphics[width=.75\columnwidth]{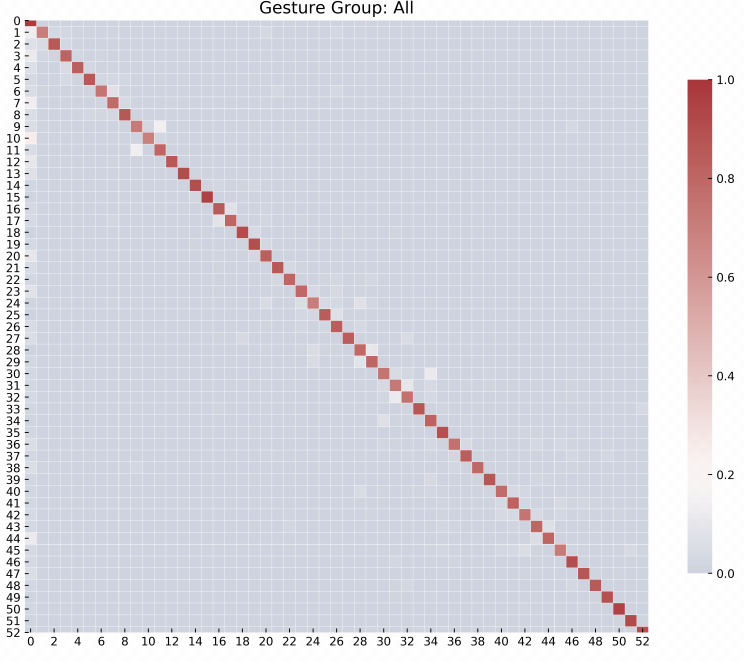}
    \caption{Confusion matrix of our results.  }
    \label{fig:lol}
    \vspace{-10pt}
\end{figure}
\begin{figure*}[t]
    \centering
    \includegraphics[width=1.5\columnwidth]{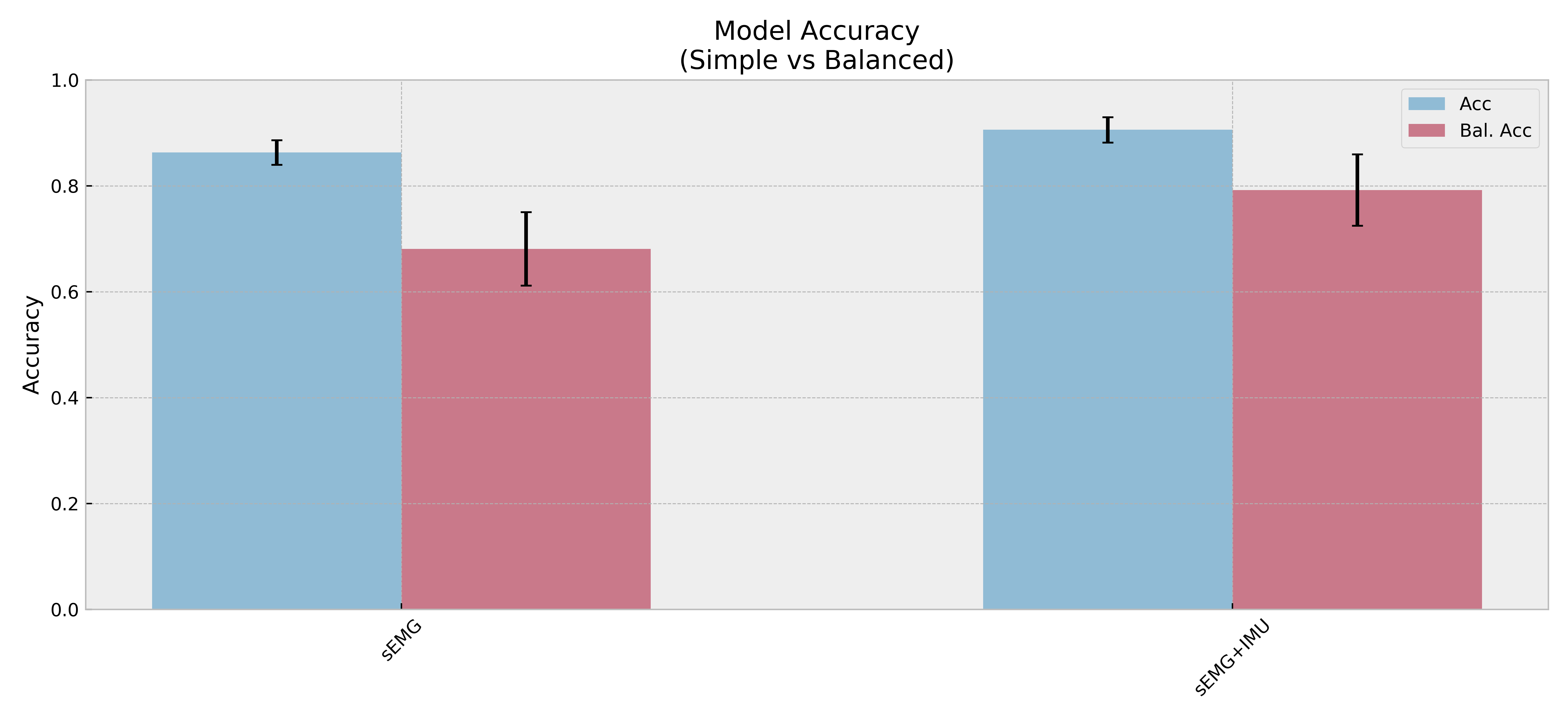}
    \vspace{-20pt}
    \caption{Accuracy calculated using inter-session cross-validation, using 5 repetitions to train and one to test, for each repetition. We see fairly consistent results across time.}
    \label{fig:eb}
    \vspace{-20pt}
\end{figure*}
\begin{table}[h!]
\begin{center}
\caption{Comparison to the current state-of-the-art. Items in \textbf{bold} mark benchmark-leading results. An asterisk(*) denotes our contribution. Each model was trained and evaluated 30 times with different seeds, with consistent results (95\% confidence interval: $\pm 0.0004$). Accuracy is used for comparison with previous work.}

\label{comptab}
\resizebox{\columnwidth}{!}{\begin{tabular}{lc}
\toprule
Model &  Accuracy             \\ \midrule
\multicolumn{2}{c}{\textit{DB5, all gestures}}\\
\textbf{Attention model with IMU*} &  \textbf{91\%}    \\
\textbf{Attention model*} &  \textbf{87\%}    \\
\citet{shen19}   & 75\% \\
\multicolumn{2}{c}{\textit{ DB5, wrist and functional gestures }}\\
\textbf{Attention model with IMU*} &  \textbf{92\%}  \\
\citet{mview}  with IMU & 91\% \\
\multicolumn{2}{c}{\textit{ DB5, just wrist gestures, no IMU}}\\
\textbf{Attention model*} & \textbf{89\%}  \\
\citet{chen2020hand} & 67\% \\
\citet{wurnn} & 62\%\\
\multicolumn{2}{c}{ \textit{DB4, all gestures}}\\
\textbf{Attention model*} & \textbf{73\%}  \\
\citet{nina5} & 69\% \\
\citet{mview} & 60\%                 \\ 
\bottomrule
\end{tabular}}
\vspace{-30pt}
\end{center}
\end{table}
An accuracy of 87\% (balanced accuracy 69\%) was achieved on NinaPro DB5, across all 53 classes.  By including the recorded IMU(accelerometer) data, the accuracy increased to 91\% (balanced accuracy 74\%). This can be seen in \autoref{fig:eb}, with error bars computed using cross-repetition cross validation. In \autoref{comptab}, we also see a benchmark-leading 73\% accuracy on DB4, a high-frequency analog to DB5. All models are implemented in Tensorflow and Numpy. We also provide a confusion matrix in \autoref{fig:lol}, which shows robust results and intuitive modes of failure (misclassifying similar gestures, for example a prismatic pinch being classified as a tripod grasp).

%


In \autoref{comptab}, we see how our simple attention-based model compares with the state-of-the-art in sEMG classification. Although our model is relatively naive and straightforward, our results are benchmark leading. As we have noted elsewhere, we see that even a simple attention-based model encoding temporal and spatial features into a compressed representation connected to a softmax layer can perform surprisingly well (79\%, as seen in \autoref{tab:bottom}).  This alone represents a significant result. 
We were able to increase accuracy gains with a distinct uptick of 8\% by training a fully connected classifier network.  This indicates that while the representation of data learned in attention is useful, there is still value and potential in using neural networks trained to work in  this newly embedded space. 

\begin{table*}[t]
\caption{Reference values for metrics. The weighted random and unweighted random values are an average of 1000 trials.}
\label{reftab}
\centering
\begin{tabular}{lcccc}
\toprule
{} &  Weighted Random &  Unweighted Random &  All Zeros &  All Ones \\
\midrule
Accuracy          &         0.406176 &           0.018954 &      0.635 &     0.009 \\
Balanced Accuracy &         0.018953 &           0.018876 &      0.019 &     0.019 \\
MCC          &         0.000028 &          -0.000006 &      0.000 &     0.000 \\
\bottomrule
\end{tabular}
\end{table*}
\begin{table*}[t]
\caption{Layer-by-layer ablative study of the network}
\label{tab:bottom}
\centering
\begin{tabular}{lcccc} 
\toprule
Model                        & No. Parameters      & Accuracy & Balanced Accuracy & MCC   \\
\multicolumn{5}{c}{\textit{Experiments on the expansion layer} }                                   \\
Fully connected              & $1.44 * 10^6$  & 87.17\%  & 69.83\%           & 0.78  \\
Conv1D                       & $1.44 * 10^6$  & 87.15\%  & 69.70\%           & 0.78  \\
Frozen fully connected layer & $1.43 * 10^6 $ & 84.18\%  & 62.90\%           & 0.73  \\
No expansion                 & $1.37 * 10^6$  & 77.80\%  & 41.79\%           & 0.59  \\
\multicolumn{5}{c}{\textit{Experiments on the attention layer }}                                   \\
Raffel attention             & $1.44*10^6$    & 86.82\%  & 69.29\%           & 0.77  \\
Temporal sum                 & $1.44*10^6$    & 86.40\%  & 67.82\%           & 0.77  \\
\multicolumn{5}{c}{\textit{Experiments on the classifier network }    }                            \\
Small classifier net         & $ 1.0 * 10^5$  & 79.79\%  & 47.81\%           & 0.65  \\
No classifier net            & $1.4 * 10 ^ 4$ & 79.71\%  & 46.58\%           & 0.64  \\
\multicolumn{5}{c}{\textit{Experiments on minor adaptations }}                                     \\
No layer normalization       & $1.43*10^6$    & 86.94\%  & 69.38\%           & 0.78  \\
Relu instead of Mish         & $1.44*10^6$    & 86.60\%  & 64.08\%           & 0.77  \\
\bottomrule
\end{tabular}
\end{table*}

\section{Discussion}

In order to better understand and clarify the source of our results, we performed an exhaustive layer by layer analysis of the proposed model. We used three metrics to assess these experiments: accuracy, class-balanced accuracy, and Matthews Correlation Coefficient (MCC) \citep{mcc, mcc2}. The MCC, formerly known as Pearson's Phi, represents the correlation coefficient between true and predicted labels, using a confusion matrix. In the case of imbalanced data, it has the useful property of going to zero when a class is completely missed. Given the imbalances typically present in laboratory-collected sEMG data, we believe the MCC is a more appropriate metric for this data.

In \autoref{reftab}, we have provided four naive baseline models to provide more context to our results. These consist of two random models, one with equally weighted class distribution and one with class distribution reflected in NinaPro DB5, and two homogeneous models. For the homogeneous models, we chose zero and one to demonstrate what expected results would be if all inputs were classified as the most common class, rest, and contrast those results with a model that classified all with one of the other non-rest classes.

We first try to understand the role of the expansion layer. In the first section of \autoref{tab:bottom}, we conduct four experiments: first, a fully connected layer is used for the expansion layer; second, 1-dimensional convolutions with kernel size 3; third, a fully connected but frozen layer; and fourth, the raw data is fed directly into the attention layer. We find that there is no significant difference between the convolutional and fully connected expansion layers, but the presence of the expansion layer does have an effect. This suggests that the allocation of space is important and allows the network to learn a more efficient representation. We also believe this suggests that the temporal properties of the data are more important than the spacial properties, as the fully connected layer assumes independence across time while the attention layer assumes independence across channels.

In the second panel, we see that a slight decrease in accuracy occurs by replacing the attention mechanism with the two alternatives proposed in \citet{raffel2015feedforward}; that mentioned in \autoref{rafeq}, and a sum across time. Although the accuracies appear similar, the sum across time approximation of attention yielded a noticeably lower class-balanced accuracy. This is consistent with the results of \citet{hopfield}, which suggest that attention can be approximated with a sum in many cases. 



In the third panel, we investigate the influence of the classifier network, with two more experiments. First, we completely remove the classifier network, feeding the output of attention into a single softmax layer. We then add a small classifier network, consisting of a single layer with 500 nodes. 
We see that without a classifier network, we still  retain about 91\% of our accuracy.
\autoref{reftab} shows that this is not due to chance, but represents a powerful result. By adding a single 500 node classifier network we don't see a increase in accuracy but even with a minimal increase in neural network depth to three layers we find that our accuracy increases.  This suggests that we are learning to encode spatiotemporal features of the data into a low-dimensional representation with our attention mechanism, and the classifier network is disentangling that representation.

Finally, in the last panel, we examine the roles of layer normalization and the Mish activation function. These did have a rather minor effect on overall accuracy but useful net positive effect on our balanced accuracy.

All models and experiments were trained using the exact same hyperparameters, mentioned in \autoref{mod}, in order to avoid any confounding factors. 

\section{Conclusion}
Our algorithm yields a benchmark leading result from a comparatively lightweight and easy to train neural network. 
Rather than starting with a strong set of priors about how the channels, time, and signal will interact, we expand the channels in order to provide sufficient space to store information and then condense using an attention layer onto a lower dimensional representation.  
That this representation only requires a simple three layer feed-forward neural network, implies that the bulk of the computational heavy lifting was achieved in the attention layer.  
One implication is that the first complexity of sEMG analysis is learning a suitable representation space. 
An intuitive way of understanding this is that sEMG sensors are not surgically attached, and therefore sensors shift from individual to individual.
Addressing this simplifies the computational needs of the final neural network.
The implication is that the attention layer is finding an invariant representation with respect to the targeted 53 gestures, as evidenced by the performance of the classifier-free model. Our robust results seems to indicate that not only does attention represent a promising avenue for future research in the electrophysiological signal space, but also that sEMG is strongly deserving of further research by the machine learning community. Our straightforward model's success in learning a robust lower-dimensional representation of sEMG offers hope that sEMG data can be used to address a wider range of complex problems, such as early diagnosis of or mediated robotic control for neurodegenerative diseases, or therapeutic tools for major injury recovery.

\bibliography{main}
\appendix
\section{SNR Augmentation} \label{snr}

\begin{figure}[H]
    \centering
    \includegraphics[width=0.7\columnwidth]{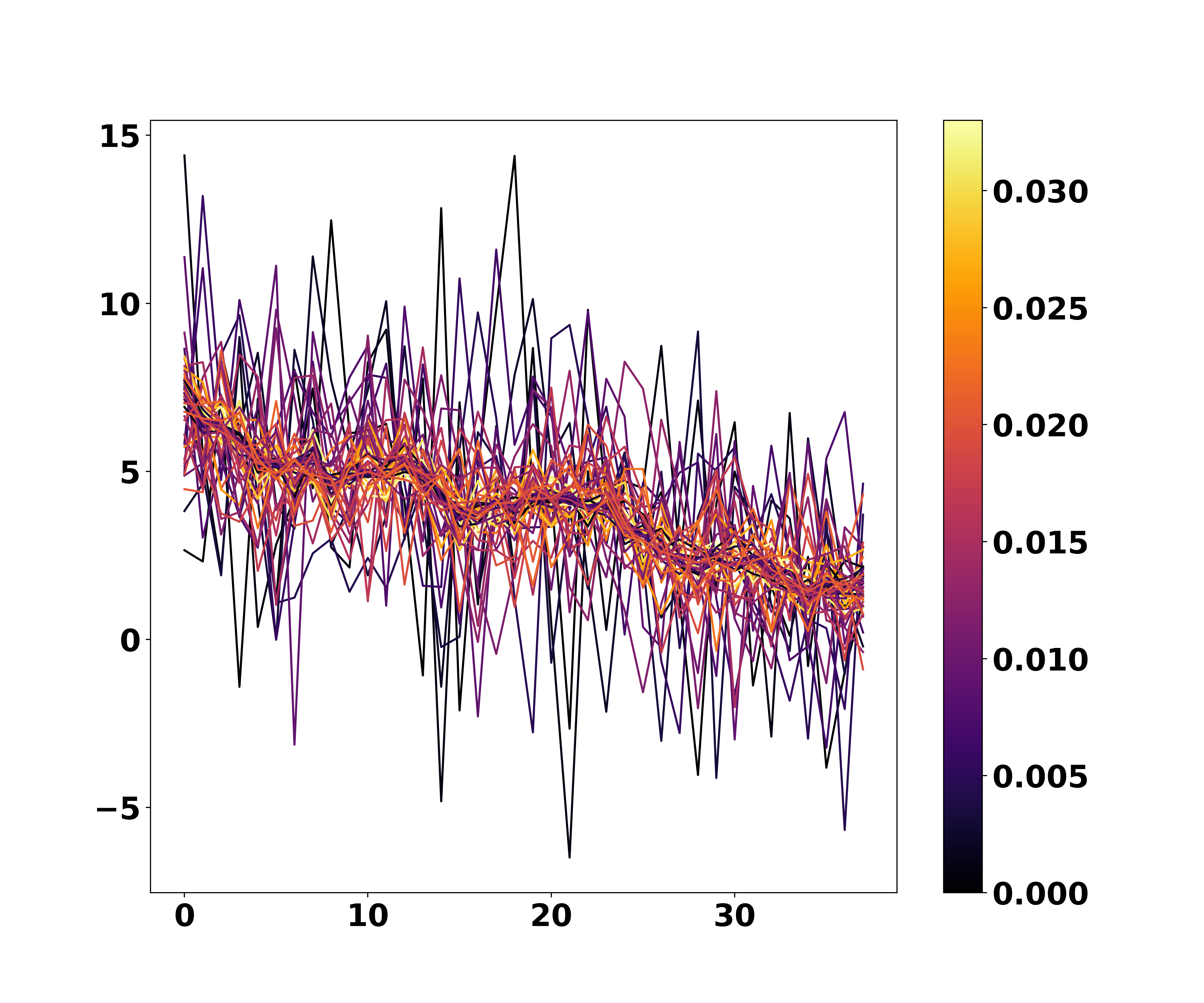}
    \vspace{-20pt}
    \caption{One channel of SNR augmentation. Brightness corresponds to probability.}
    \label{fig:my_label}
\end{figure}
The signal to noise augmentation is performed in the following manner. First, the signal power $P_s$ in decibels of signal $S\,\epsilon \,\mathbb{R}^{T \times C}$ calculated as follows:

\begin{equation}
    P_s = 10 \mathrm{log}_{10} \left( \frac{\sum_{t=1}^{T} S_t^2}{T} \right)
\end{equation}
Then, we calculate the power of the noise in decibels required to reach the appropriate desired SNR on all channels:
\begin{equation}
    P_n = \mathrm{SNR} - P_s
\end{equation}
We then convert the noise back out of decibels and take the square root to calculate the standard deviation of the noise to be added to each channel:
\begin{equation}
    \sigma_n = \sqrt{10 ^ {\frac{P_n }{ 10}}}
\end{equation}
Finally, we calculate the noise $w\, \epsilon \,\mathbb{R}^{T \times C}$:
\begin{equation}
    w \sim \mathcal{N}(0,\,\sigma_n^{2})\,
\end{equation}
This noise is then simply added to the existing signal:
\begin{equation}
S_{\mathrm{augmented}} = S + w
\end{equation}
The likelihood of a SNR to be selected during augmentation directly proportional to the target SNR's value, for example an SNR of 30 is twice as likely to occur as an SNR of 15, and so forth. An example augmentation can be seen in \autoref{fig:my_label}.
\end{document}